\documentclass[12pt]{article}
\usepackage{amssymb}
\usepackage{amsmath,amsthm}
\usepackage{amsfonts}
\usepackage{graphicx}
\usepackage{graphics}
\usepackage{xcolor}
\usepackage{algorithm}
\usepackage{algpseudocode}

\numberwithin{equation}{section}
\numberwithin{footnote}{section}

\newcommand{\bp}{\begin{proof}}
\newcommand{\ep}{\end{proof}}
\newcommand{\be}{\begin{equation}}
\newcommand{\ee}{\end{equation}}
\newcommand{\bes}{\begin{equation*}}
\newcommand{\ees}{\end{equation*}}

\begin{document}
\date{\small\textsl{\today}}
\title{
The use of the symmetric finite difference in the \\ local binary pattern (symmetric LBP)
}
\author{Zeinab Sedaghatjoo
\begin{footnote}{
Corresponding author.\newline {\em  E-mail addresses:}
\newline z.sedaqatjoo@aut.ac.ir , zeinab.sedaghatjoo@gmail.com (Z. Sedaghatjoo) \\
h{\_}hosseinzadeh@aut.ac.ir  ,  hosseinzadeh@pgu.ac.ir (H. Hosseinzadeh). 
}$\vspace{.2cm} $
\end{footnote} 
, Hossein Hosseinzadeh
\\
\small{\em  Department of Mathematics, Persian Gulf University, Bushehr, Iran.}
\vspace{-1mm}} \maketitle
%-------------------------------------------------------------
\vspace{.9cm}
%%%%%%%%%%%%%%%%%%%%%%%%%%%%%%%%%%%%%%%%%%%%%%%%%%%%%%%%%%%%%%%%%%%%%%%%%%%%%%%%%%%%%%%%%%%%%%%%
 
\begin{abstract}
The paper provides a mathematical view to the binary numbers presented in the Local Binary Pattern (LBP) feature extraction process.
Symmetric finite difference is often applied in numerical analysis to enhance the accuracy of approximations.
Then, the paper investigates utilization of the symmetric finite difference in the LBP formulation for face detection and facial expression recognition. 
It introduces a novel approach that extends the standard LBP, which typically employs eight directional derivatives, to incorporate only  four directional derivatives. 
This approach is named symmetric LBP.  
The number of LBP features is reduced to 16 from 256 by the use of the symmetric LBP. 
The study underscores the significance of the number of directions considered in the new approach. 
Consequently, the results obtained emphasize the importance of the research topic.
\vspace{.5cm}\\
\textbf{{\em Keywords}}: 
{local binary pattern, 
feature extraction,
face detection,
facial expression recognition, 
symmetric finite difference.} \\
MSC 2020:  {68W01, 65A05, 65L12}%
\end{abstract}

\section{Introduction}
Local Binary Pattern (LBP) is a texture descriptor used in computer vision for image classification. It was first introduced by Ojala et al. \cite{ojala1996comparative} and has gained popularity due to its effectiveness in extracting texture features while maintaining computational simplicity. The core concept of the LBP involves comparing each pixel of an image with its neighbouring pixels to encode the local texture information into binary patterns. In mathematical sense, the comparison test can be interpreted as directional derivatives. Non negative derivations conclude 1 and negative ones lead $0$ in this stage.  The resulting binary values obtained from the comparisons are concatenated sequentially in a clockwise order, forming an 8-digit binary number for each pixel. A histogram is computed over the entire pixels, capturing the frequency of different LBPs. This histogram serves as a feature vector for the image. Then, 256 LBPs are reduced to 16 or less features by the histogram. Since the extracted features mainly describe the entire image, the image is divided into several regions to capture more local features. This is achieved by deriving histograms from the LBPs extracted from the regions and concatenating them. We show in this paper that the number of LBPs can be reduced to 16 instead of 256 if one uses symmetric  directional derivatives. In this case, there is no need to use the histogram in the LBP. So, the computational cost of the LBP process is reduced.

Variations of LBP have been developed to enhance its performance and address specific challenges in different applications \cite{pajdla2004computer,huo2019effective, kaplan2020brain, vu2022masked}. Some notable variations of the LBP are presented, such as: Uniform Local Binary Patterns (ULBP), Rotation-Invariant Local Binary Patterns (RILBP), Local Ternary Patterns (LTP), Multi-scale Local Binary Patterns (MLBP) and Completed Local Binary Patterns (CLBP). A complete review of LBP developments is presented in \cite{khaleefah2020review}.
Experimental results presented in \cite{zhang2009local} show that LBP is significantly enriched if more directional derivatives are applied in the formulations. Additionally, reference \cite{martolia2020modified} claimed that four directional derivatives can extract texture patterns of an image as effectively as eight directional derivatives. This suggests that the 8-bit binary numbers used in traditional LBP can be replaced by 4-bit representations. This led the authors to further investigate the role of directional derivatives and their approximation using the symmetric finite difference. The findings confirm the claims made in regarding the effectiveness of using four directional derivatives for texture analysis if the directional derivatives are approximated by the symmetric finite difference. %The new proposed approach is named symmetric LBP.
%Note that in \cite{tabatabaei2015facial} 

Then, in this work, a new approach, named symmetric LBP, is proposed. This approach led to more accurate approximations of the directional derivatives in the LBP. Furthermore, 4-bit binary numbers are found from this approach which are as effective as the traditional 8-bit representations. This leads less computational cost. The efficiency of the new 4-bit LBP descriptor are verified through experiments on face detection and facial expression recognition tasks.
Forthcoming sections of the paper are designed as follows:
\begin{itemize}
\item
In Section \ref{sec2}, the LBP is described.
\item
In Section \ref{sec3}, the directional derivatives of the LBP are studied and the symmetric LBP is proposed there to enhance the standard LBP.
\item
In Section \ref{sec4}, some experiment results are presented in face detection and facial expression recognition.
\item
Finally, the article is completed by a brief conclusion in Section \ref{sec5}.
\end{itemize}

%%%%%%%%%%%%%%%%%%%%%%%%%%%%%%%%
\section{Local Binary Pattern} \label{sec2}
\begin{figure}[h!]
\centering
{\includegraphics[scale=0.2]{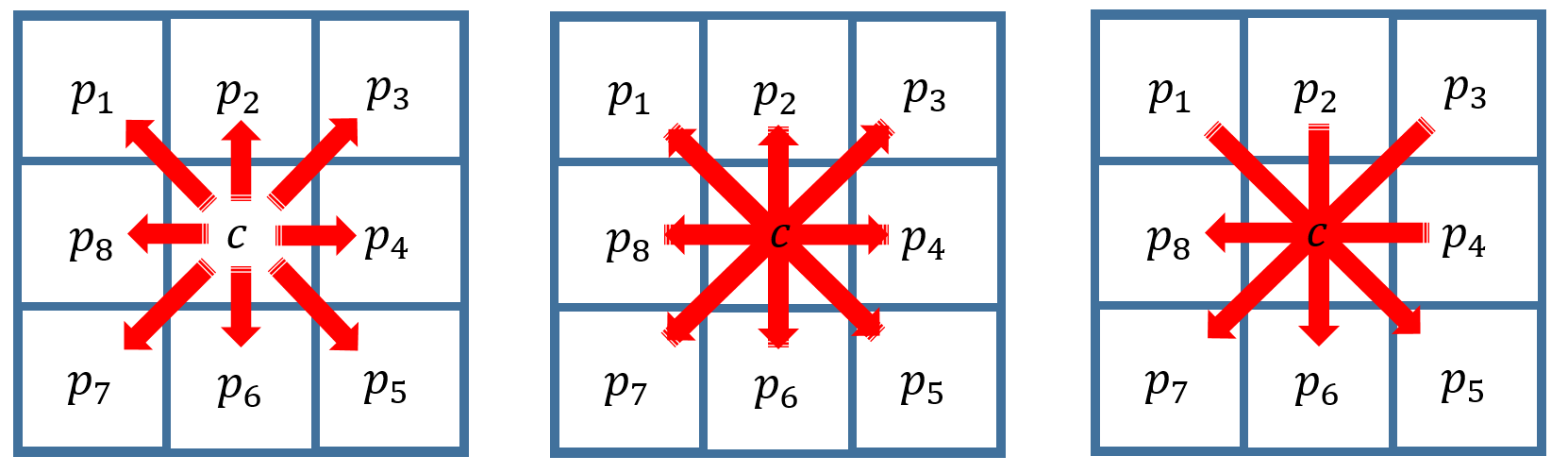}} \\
~~~~~~~~(A)~~~~~~~~~~~~~~~~~(B)~~~~~~~~~~~~~~~~~~(C)~~~~~~~~ \\
\caption{Directional derivatives respect to the standard LBP (A), 8-bit symmetric LBP (B) and 4-bit symmetric LBP (C).}\label{im1}
\end{figure}

LBP is a texture descriptor. It operates on images by labelling the pixels via comparing their values with the values of their neighbours \cite{pajdla2004computer}. For applying the LBP, an image is divided into some local regions and the LBP texture descriptors are extracted from each region, successive. The descriptors are then concatenated to form a global description of the image. The LBP descriptor takes 3×3-pixel blocks as the basic unit, and the difference between the central pixel and its adjacent 8 pixels is extracted as local texture feature representation. It can be formulated as follows:
\begin{equation}\label{s}
S(p,c)=
\left \{
\begin{array}{ll}
1, ~~~~~~~~~~~ \hbox{if} ~~ g(p) \geq g(c),\\
\\
0, ~~~~~~~~~~~ \hbox{if} ~~ g(p) < g(c).
\end{array}\right.
\end{equation}
where $g(p)$ and $g(c)$ are the value of the neighbouring pixel $p$, and the central pixel $c$, respectively. Consequentially, if we label the neighbouring pixels as $p_1, p_2, ..., p_8$ same as Figure \ref{im1}, an 8-digit binary number will be encoded to a LBP as:
\begin{equation}\label{lbp}
lbp=\sum_{i=1}^{8} 2^{i-1} S(p_i,c).
\end{equation}
Note that, equations \eqref{s} and \eqref{lbp} can be described based on directional derivatives $d_i$, defined as
\[
d_i=\lim_{p_i \rightarrow c} \frac{g(p_i)-g(c)}{ \| p_i-c \|},
\] 
for $i=1, 2, ..., 8$, and $Hardlim$ function $H$ as 
\begin{equation}\label{lbp2}
lbp=\sum_{i=1}^{8} 2^{i-1} H(d_i),
\end{equation}
where $H$ refers $0$ and $1$ for negative and non-negative numbers, respectively. Here, $d_i$ is derivative of two dimensional function $g$ in direction $v_i=p_i-c$ at center point $c$. Direction $v_i$ are shown in  \ref{im1} (A) for $i=1, 2, ..., 8$. 
After obtaining the LBP values for the pixels of the image, a histogram is constructed to represent the frequency of occurrence of different LBPs within the image. This histogram effectively summarizes the local texture patterns present in the image. Since there are $256$ patterns in the histogram, we mostly summarize it in $8$ to $16$ patterns \cite{pajdla2004computer, khaleefah2020review}. 

%%%%%%%%%%%%%%%%%%%%%%%%%%%%%%%%%%%%%%%%%%%%%%%%%%%%%%
\section{Directional derivatives and the symmetric LBP} \label{sec3}
Finite difference formulas are numerical techniques used to approximate directional derivatives of functions when analytical methods are not feasible or when dealing with discrete data points. Directional derivatives represent the rate of change of a function in a specific direction, providing valuable insights into how the function varies along a given path. 
One common approach to approximating directional derivatives is through one-sided finite differences. These formulas involve computing the difference in function values at two nearby points along the desired direction and dividing by the step size. The LBP uses these concept by applying
\begin{equation}\label{old}
d_i \simeq {g(p_i)-g(c)} , ~~~ i=1, 2, ... ,8 ,
\end{equation}
and considering ${\| p_i-c \| }=1$. These approximations lead the accuracy of order $1$ \cite{leveque2007finite}.
One common approach for approximating the directional derivatives more accurate is applying central differencing with an error of order $2$ \cite{leveque2007finite}. The central differencing involves taking one forward Taylor step and one backward Taylor step, then subtracting these steps. Then, the directional derivative formula can be calculated by the central differencing as:
\begin{equation}\label{new}
d_i \simeq \frac{g(p_i)-g(\bar{p_i})}{2}  , ~~~ i=1, 2, ... ,8 ,
\end{equation}
where $\bar{p_i}$ is that pixel in apposite to $p_i$. In mathematical sense, $\bar{p}_i=2c-p_i$ if the points are considered as vectors with two coordinates. In Figure \ref{im1} (B) the engaged pixels in to the new approximated derivatives are shown. From this figure $\bar{p}_1=p_5, \, \bar{p}_2=p_6, \bar{p}_3=p_7,  \,\bar{p}_4=p_8,  \,\bar{p}_5=p_1,  \,\bar{p}_6=p_2,  \,\bar{p}_7=p_3$ and $\bar{p}_8=p_4 $.  The formula, Equation (\ref{new}), also is named symmetric finite difference in the numerical analysis \cite{leveque2007finite}.

\begin{figure}[h!]
\centering 
{\includegraphics[scale=2]{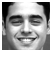}} {\includegraphics[scale=0.3]{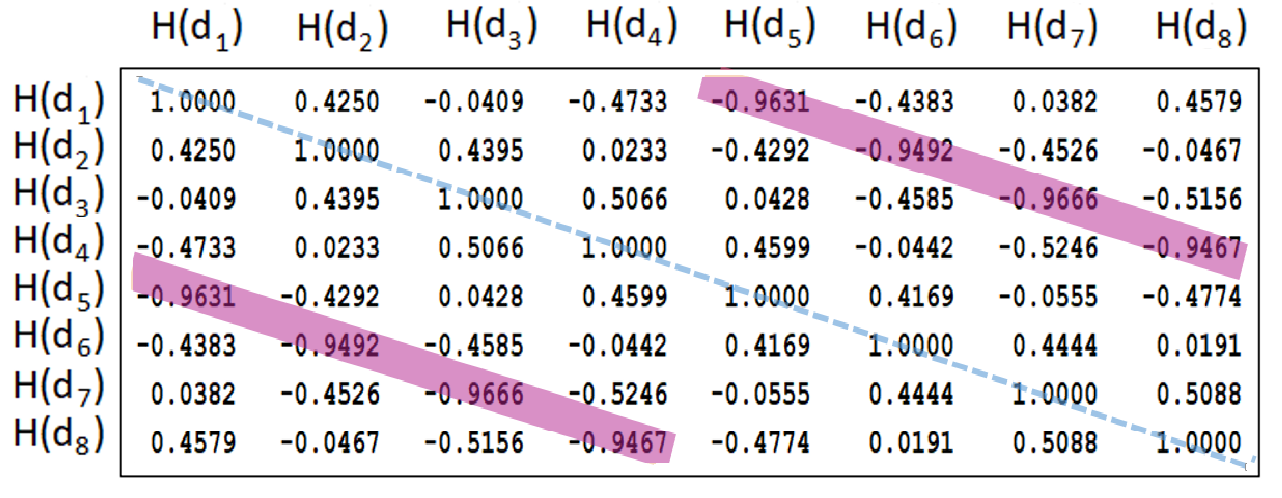}}
\caption{The correlation matrix showing the correlation coefficients between variables $H(d_i)$ for a face image. The coefficients between $H(d_i) $ and $H(d_{i+4})$ is less than $-0.94$ highlights the linear dependency between the variables. }\label{corr}
\end{figure}

If we use new approximations \eqref{new} in the LBP formulation, we see $d_i=-d_{i+4}$ for $i=1, 2, 3$ and $4$. So, the values of $d_i$ and $d_{i+4}$ are linearly dependent. It leads, $H(d_i) $ and $H(d_{i+4})$ are mostly correlated. 
In mathematical sense, one can see
\begin{align}
H(d_i) = 1-H(d_{i+4}) ,
\end{align}
when $d_i \neq 0$. This dependency may reduce the efficiency of the LBP. Correlation matrix showing the correlation coefficients between variables $H(d_i)$ is presented in Figure \ref{corr} for a face image. The correlation coefficient measures how much and in which direction two variables are linked linearly \cite{schott2016matrix}. From Figure \ref{corr}, the correlation coefficients between $H(d_i) $ and $H(d_{i+4})$ is around $-0.95$ which it highlights the linear dependency between the variables. This dependency does not occur with the same intensity for the standard LBP, which applies symmetric approximation (\ref{old}). The correlation coefficient between the variables $H(d_i)$ and $H(d_{i+4})$ in the standard LBP is around $-0.65$. 
%%%%%%%%%%
The linear dependency between variables may reduce the LBP efficiency as is shown in the next section. To remove this drawback, one can ignore four first directional derivatives, $d_{1}, d_{2}, d_{3}$ and $d_{4}$, and use only the last four directional derivatives, $d_{5}, d_{6}, d_{7}$ and $d_{8}$. The directions of the selected derivatives are shown in Figure \ref{im1} (C). Then a new LBP formulation, obtained based on the extracted directional derivatives, is
\begin{align}\label{lbp3}
lbp & =\sum_{i=1}^{4} 2^{i-1} H(d_{i+4}) .
\end{align}
Therefore, three versions of the LBP may be applied for feature extraction:
\begin{itemize}
\item
 Standard LBP: using eight directional derivatives approximated by Equation (\ref{old}). This version is referred as StLBP here.
\item
 8-bit symmetric LBP: using eight directional derivatives approximated by Equation (\ref{new}). This version is referred as SyLBP8 here.
\item
4-bit symmetric LBP: using four directional derivatives, $d_5-d_8$, approximated by Equation (\ref{new}). This version is referred as SyLBP4 here.
\end{itemize}

\begin{figure}[h!]
\centering
{\includegraphics[scale=0.3]{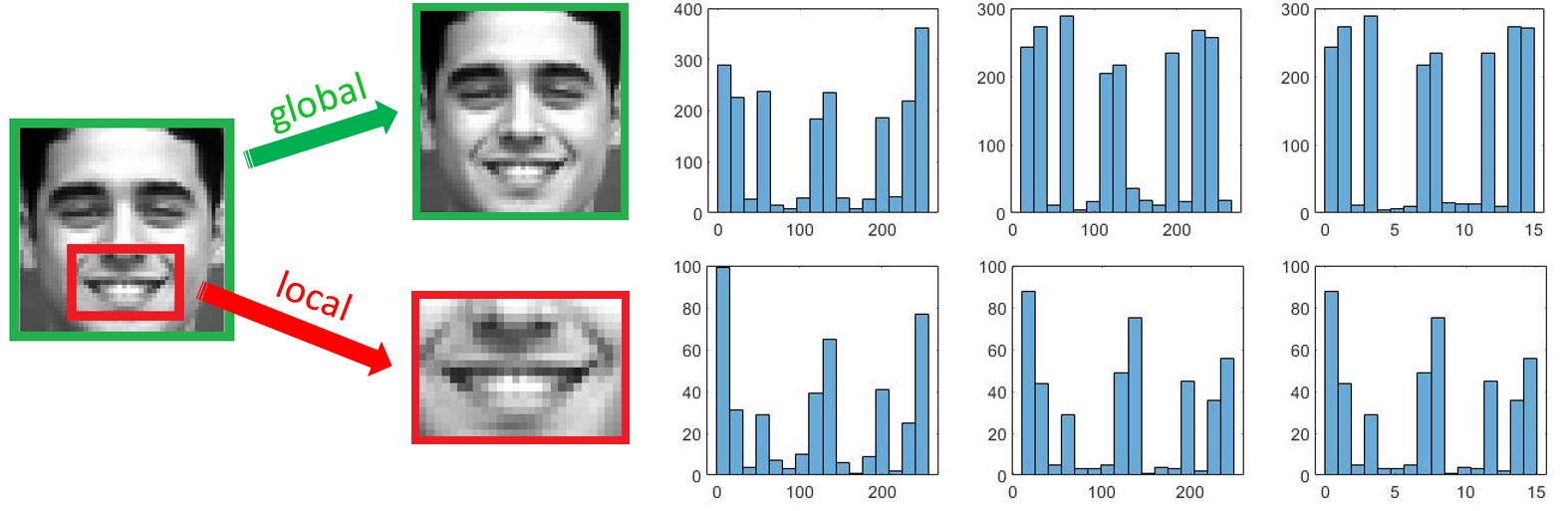}}
\caption{The histogram respect to StLBP, SyLBP8 and SyLBP4 within a global and a local images is presented from left to right, respectively.}\label{im2}
\end{figure}

The Histogram of frequency of occurrence of different LBP patterns within two different images
respecting StLBP, SyLBP8 and SyLBP4 are presented in Figure \ref{im2}. From Figure \ref{im2}, the histograms of
SyLBP8 and SyLBP4 are similar for a face image but they are slightly different for the mouth image. 
%Then the features extracted by the SyLBP4 are similar to the features obtained by the SyLBP8.
However, they are different from those of StLBP. 
Note, in the StLBP and SyLBP8, 256 patterns are extracted which are then reduced to 16 features using the histogram, while the SyLBP4 extracts 16 patterns directly.

%%%%%%%%%%%%%%%%%%%%%%%%%%%%%%%%%%%%%%%%%%%%%%%%%%%%%%
\section{Experiment results and discussions}\label{sec4}
\begin{figure}[h!]
\centering
{\includegraphics[scale=0.3]{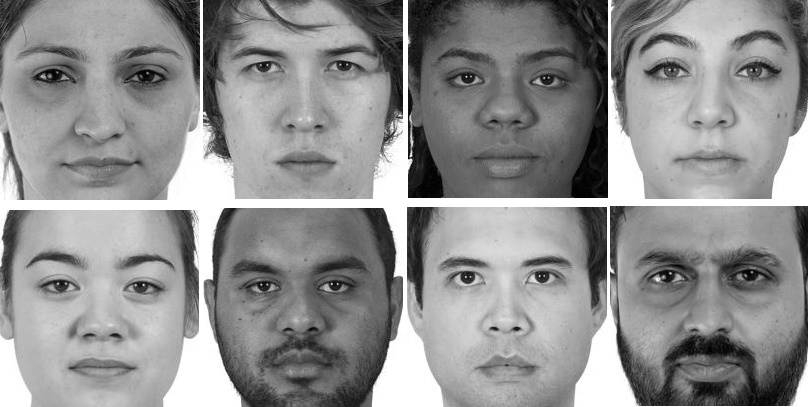}} ~~~
{\includegraphics[scale=0.3]{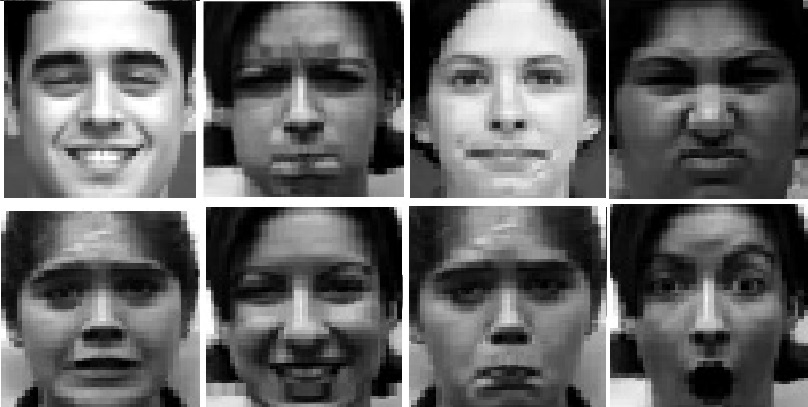}}
\caption{Some human faces for face detection (left) and facial expression recognition (right). These images are presented in CFD, CFD-MR, CFD-INDIA and CK datasets. }\label{faceim}
\end{figure}

In this section, we present the outcomes of our experiments for face detection and facial expressions recognition. We use databases CFD \cite{ma2015chicago}, CFD-MR \cite{ma2021chicago},
 and CFD-INDIA \cite{lakshmi2021india},  containing 1410 face and 1410 clutter images for face detection. The facial regions were manually
 cropped and resized to images of size $64\times64$. And CK dataset \cite{lucey2010extended} is applied for  facial expressions classification. The CK dataset comprises 981 gray-scale images of facial expressions, each measuring $48 \times 48$ pixels. These images are annotated into seven distinct emotional classes: anger (135 samples), contempt (54 samples), disgust (177 samples), fear (75 samples), happiness (207 samples), sadness (84 samples), and surprise (249 samples). Prior to conducting experiments, we preprocessed the dataset by dividing each facial image into 16 sub-images. This preprocessing step facilitated the extraction of LBP features from the facial expressions. Some facial images presented in the databases are demonstrated in Figure \ref{faceim}.

For LBP feature extraction, we used the standard formulation \eqref{lbp}, second-order formulation (\ref{lbp2}) and the reduced formulation \eqref{lbp3} for SLBP, NLBP8 and NLBP4, respectively. These features were then classified by the Support Vector Machine (SVM), with 70\% of pictures allocated for training and 30\% for testing. We used both Linear SVM and Non-linear SVM for classification. For non-linear SVM classification using the radial basis function (RBF) kernel, we set the kernel parameter $\gamma = 100$ (determines the influence of individual training samples, with low values meaning 'far' and high values meaning 'close') and the regularization parameter $C = 10$ (balances the classification error term and the regularization term. A smaller $C$ allows for a larger margin and may result in more miss classifications but a simpler model, while a larger $C$ emphasizes classifying the training data correctly and may result in a smaller margin but potentially a better fit to the training data).

Table \ref{table1} presents the accuracy of the face detection classification,
and Table \ref{table2} presents the accuracy of each class obtained through our experiments in  facial expressions recognition.
From the results presented, it can be observed that the standard LBP (StLBP) is more accurate than the 8-bit symmetric LBP (SyLBP8) and nearly as accurate as the 4-bit symmetric LBP (SyLBP4). This suggests that the 4-bit symmetric LBP is able to effectively extract the relevant texture patterns in the datasets, while the 8-bit symmetric LBP does not provide this. The reason for this is that the 4-bit symmetric LBP only extracts 16 patterns and does not require a histogram-based dimension reduction, unlike the 8-bit symmetric LBP which extracts 256 patterns and relies on the histogram process. Since the histogram-based dimension reduction may not be perfect, the 8-bit symmetric LBP does not necessarily present more efficient features compared to the 4-bit version. Therefore, the authors suggest applying the 4-bit symmetric LBP instead of the standard LBP, as it leads to only 16 patterns instead of 256 while maintaining accurate results. This provides a more efficient and compact representation of the texture information in the images.

\begin{table}[h!]\label{table1}
\caption{Accuracy of the SVM for face detection. Three different LBPs are applied for feature extraction.}\label{table1}
\begin{footnotesize}
\begin{tabular}{l*{7}{c}r}
\hline
\hline
               &   &   &  & Accuracy ($\%$)   &  &  &  \\
\cline { 2 - 8}
                    &   & Linear SVM&   &   &   &  RBF  SVM &  \\
\cline { 2 - 4} \cline { 6 - 8}

LBP       &  StLBP &  SyLBP8 &  SyLBP4 &  &  StLBP & SyLBP8 & SyLBP4 
$\vspace{.1cm} $ 
\\
face     & 99.99  & 99.96 &  99.94 &        & 99.76 &  99.27 &  99.44  \\
clutter     & 99.30 &  99.21 &  99.07 &        & 100 &  99.98 &  100  \\
\hline
total     & 99.75  & 99.58  & 99.50 &        &   99.88 &  99.63 &  99.77  \\
\hline
\hline

\end{tabular}
\end{footnotesize}
\end{table}
%%%%%%%%%%%%%%%%%%

\begin{table}[h!]\label{table2}
\caption{Accuracy of the SVM for face emotional recognition. Three different LBPs are applied for feature extraction.}\label{table2}
\begin{footnotesize}
\begin{tabular}{l*{7}{c}r}
\hline
\hline
               &   &   &  & Accuracy ($\%$)   &  &  &  \\
\cline { 2 - 8}
                    &   & Linear SVM&   &   &   &  RBF  SVM &  \\
\cline { 2 - 4} \cline { 6 - 8}

LBP       &  StLBP &  SyLBP8 &  SyLBP4 &  &  StLBP & SyLBP8 & SyLBP4 
$\vspace{.1cm} $ 
\\
anger     & 95.7  & 90.6 &  92.2 &        & 82.7 &  84.0 &  87.3  \\
contempt     & 90.6 &  81.2 &  86.8 &        & 87.6 &  78.7 &  86.1  \\
disgust    & 95.7 &  92.4 &  94.6 &         & 88.7 &  83.4 &  88.4  \\
fear    & 95.6 &  91.9 &  94.6 &         & 92.1 &  85.4 &  89.0  \\
happiness    & 97.4 &  96.6 &  98.0 &         & 97.2 &  95.6 &  97.3  \\
sadness    & 92.8 &  88.5 &  93.5 &         & 87.5 &  81.0 &  86.5  \\
surprise    & 97.4 &  96.3 &  97.4 &         & 96.9 &  94.8 &  96.5  \\
\hline
total     & 96.0  & 93.0  & 95.2 &        &   91.8 &  88.6 &  92.0  \\
\hline
\hline
\end{tabular}
\end{footnotesize}
\end{table}
%%%%%%%%%%%%%%%%%%

%%%%%%%%%%%%%%%%%%%%%%%%%%%%%%%%%%%%%%%%%%%%%%%%%%%%%%
\section{Conclusion}\label{sec5}
In this study, we conducted experiments on the CFD, CFD-MR, CFD-INDIA and CK datasets to explore the classification of the face detection and facial expressions recognition using Local Binary Pattern (LBP) features and Support Vector Machine (SVM) models. Our primary objective was to evaluate the effectiveness of using directional derivatives in the LBP feature extraction and the accuracy of the SVM classification.
We proposed the symmetric LBP and we found using 8-bit binary numbers is not necessary and one can use 4-bit binary numbers when the directional derivatives  are approximated by the symmetric finite differences.
Then, the efficiency of the LBP enhanced when four
first directional derivatives, as outlined in the paper, are removed from the LBP formula and only four last ones are considered. So, the symmetric LBP, applied 4-bit binary numbers, yielded promising results in extracting discriminative features from facial images. 
Furthermore, our experiments demonstrated that SVM classifiers, both Linear and Non-linear with radial basis function kernel, were effective in classifying emotional expressions.
It is suggested to explore the implementation efficiency of the symmetric finite differences in larger pixel blocks, such as  $5 \times 5$ and  $7 \times 7$ blocks, to further evaluate their impact on LBP performance.
Looking ahead, future research could focus on applying the symmetric LBP on larger and more diverse datasets, to find the efficiency of the new approach for other problems. The idea also can be extended to use more complex directional derivative formulas with higher order of approximation. This may enhance the efficiency of the LBP.

%%%%%%%%%%%%%%%%%%%%%%%%%%%%%%%%%%%%%%%%%%%
% References

\bibliographystyle{elsarticle-num}
\bibliography{mybibfile}

\end{document}